\documentclass[12pt]{article}
\pdfoutput=1
\usepackage[utf8]{inputenc}
\usepackage[T1]{fontenc}
\usepackage{float}
\usepackage{authblk}
\usepackage[normalem]{ulem}
\usepackage{amsmath}
\usepackage{amsthm}
\usepackage{adjustbox}
\usepackage{graphicx}
\usepackage{booktabs}
\usepackage{mathtools}
\usepackage{fullpage}
\usepackage[mathlines]{lineno}
\usepackage{subcaption}
\usepackage[style=numeric-comp,sorting=none,maxnames=25]{biblatex} 
\usepackage{hanging}
\usepackage{longtable}

\usepackage{multirow}
\usepackage[table,xcdraw]{xcolor}

\usepackage[colorlinks = true,
            linkcolor = black,
            urlcolor  = black,
            citecolor = teal]{hyperref}

\newcolumntype{L}[1]{>{\raggedright\let\newline\\\arraybackslash\hspace{0pt}}m{#1}}

\title{Designing Ecosystems of Intelligence\\from First Principles}

\author[1,2]{Karl~J.~Friston}
\author[1,2]{Maxwell~J.D.~Ramstead\thanks{\href{maxwell.ramstead@verses.ai}{maxwell.ramstead@verses.ai}}}
\author[1,3]{Alex~B.~Kiefer}
\author[1]{Alexander~Tschantz}
\author[1,4]{Christopher~L.~Buckley}
\author[1,5]{Mahault~Albarracin}
\author[1,6]{Riddhi~J.~Pitliya}
\author[1,7,8,9]{Conor~Heins}
\author[1,10]{Brennan~Klein}
\author[1,11]{Beren~Millidge}
\author[1,12,13,14]{Dalton~A.R.~Sakthivadivel\thanks{\href{dalton.sakthivadivel@verses.ai}{dalton.sakthivadivel@verses.ai}}}
\author[1,6,15]{Toby~St~Clere~Smithe}
\author[1,16]{Magnus~Koudahl}
\author[1,17]{Safae~Essafi~Tremblay}
\author[1]{Capm~Petersen}
\author[1]{Kaiser~Fung}
\author[1]{Jason~G.~Fox}
\author[1]{Steven Swanson}
\author[1]{Dan Mapes}
\author[1]{Gabriel~René}

\affil[1]{VERSES AI Research Lab, Los Angeles, California, USA}
\affil[2]{Wellcome Centre for Human Neuroimaging, University College London, London, UK}
\affil[3]{Department of Philosophy, Monash University, Melbourne, Australia}
\affil[4]{Sussex AI Group, Department of Informatics, University of Sussex, Brighton, UK}
\affil[5]{Department of Computer Science, Universit\'e du Qu\'ebec \`a Montr\'eal, Montr\'eal, Qu\'ebec, Canada}
\affil[6]{Department of Experimental Psychology, University of Oxford, Oxford, UK}
\affil[7]{Department of Collective Behaviour, Max Planck Institute of Animal Behavior,\protect\\Konstanz, Germany}
\affil[8]{Department of Biology, University of Konstanz, Konstanz, Germany}
\affil[9]{Centre for the Advanced Study of Collective Behaviour, University of Konstanz,\protect\\Konstanz, Germany}
\affil[10]{Network Science Institute, Northeastern University, Boston, Massachusetts, USA}
\affil[11]{Brain Network Dynamics Unit, University of Oxford, Oxford, UK}
\affil[12]{Department of Mathematics, Stony Brook University, Stony Brook, New York, USA}
\affil[13]{Department of Physics and Astronomy, Stony Brook University, Stony Brook, New York, USA}
\affil[14]{Department of Biomedical Engineering, Stony Brook University, Stony Brook, New York, USA}
\affil[15]{Topos Institute, Berkeley, California, USA}
\affil[16]{Department of Electrical Engineering, Eindhoven University of Technology,\protect\\Eindhoven, The Netherlands}
\affil[17]{Department of Philosophy, Universit\'e du Qu\'ebec \`a Montr\'eal, Montr\'eal, Qu\'ebec, Canada}

\begin{document}
\maketitle
\pagenumbering{arabic}

\vspace{-1.0cm}
\tableofcontents
\vspace{0.75cm}
\begin{abstract}
This white paper lays out a vision of research and development in the field of artificial intelligence for the next decade (and beyond). Its denouement is a cyber-physical ecosystem of natural and synthetic sense-making, in which humans are integral participants---what we call ``shared intelligence''. This vision is premised on \emph{active inference}, a formulation of adaptive behavior that can be read as a physics of intelligence, and which inherits from the physics of self-organization. In this context, we understand intelligence as the capacity to accumulate evidence for a generative model of one's sensed world---also known as self-evidencing. Formally, this corresponds to maximizing (Bayesian) model evidence, via belief updating over several scales: i.e., inference, learning, and model selection. Operationally, this self-evidencing can be realized via (variational) message passing or belief propagation on a factor graph. Crucially, active inference foregrounds an existential imperative of intelligent systems; namely, curiosity or the resolution of uncertainty. This same imperative underwrites belief sharing in ensembles of agents, in which certain aspects (i.e., factors) of each agent's generative world model provide a common ground or frame of reference. Active inference plays a foundational role in this ecology of belief sharing---leading to a formal account of collective intelligence that rests on shared narratives and goals. We also consider the kinds of communication protocols that must be developed to enable such an ecosystem of intelligences and motivate the development of a shared hyper-spatial modeling language and transaction protocol, as a first---and key---step towards such an ecology.
\end{abstract}

\section{Introduction}\label{sec:intro}

This white paper presents active inference as an approach to research and development in the field of artificial intelligence (AI), with the aim of developing \textit{ecosystems} of natural and artificial intelligences. The path forward in AI is often presented as progressing from systems that are able to solve problems within one narrow domain---so-called ``artificial narrow intelligence'' (ANI)---to systems that are able to solve problems in a domain general-manner, at or beyond human levels: what are known as ``artificial general intelligence'' (AGI) and ``artificial super-intelligence'' (ASI), respectively \cite{bostrom2014superintelligence}. We believe that approaching ASI (or, for reasons outlined below, even AGI) likely requires an understanding of networked or collective intelligence. Given the growing ubiquity of things like autonomous vehicles, robots, and arrays of edge computing devices and sensors (collectively, the internet of things), the zenith of the AI age may end up being a distributed network of intelligent systems, which interact frictionlessly in real time, and compose into emergent forms of intelligence at superordinate scales. The nodes of such a distributed, interconnected ecosystem may then be human users as well as human-designed artifacts that embody or implement forms of intelligence.

In order to enable such ecosystems, we must learn from nature. While acknowledging neuroscience as a key inspiration for AI research, we argue that we must move beyond brains, and embrace the active and nested characteristics of natural intelligence, as it occurs in living organisms and as it might be implemented in physical systems more generally. In our view, this entails asking not only ``How does intelligence present to us, as researchers?'' but also, crucially, the complementary question ``What is it that intelligence \textit{must be}, \textit{given that} intelligent systems exist in a universe like ours?'' To address this challenge, we aim to deduce fundamental properties of intelligence from foundational considerations about the nature of persisting physical systems (i.e., ``first principles'').

In so doing, we foreground \textit{active inference}, which combines the virtues of such a first-principles, physics-based approach to AI with Bayesian formulations, thus reframing and, in some key respects, extending the methods found in Bayesian approaches to machine learning, which provide the foundations of state-of-the-art AI systems. Active inference is an account of the inevitable existence of agency in physical worlds such as ours, which motivates a definition of intelligence as the capacity of systems to generate evidence for their own existence. This encompasses cognition (i.e., problem-solving via action and perception) and curiosity, as well as the capacity for creativity, which underwrites the current interest in generative AI \cite{Sequoia2022}. Active inference offers a formal definition of intelligence for AI research, and entails an explicit mechanics of the beliefs of agents and groups of agents---known as Bayesian mechanics \cite{Ramstead2022a, DaCosta2021}---which is uniquely suited to the engineering of ecosystems of intelligence, as it allows us to write down the dynamics of sparsely coupled systems that self-organize over several scales or ``levels'' \cite{Friston2015a, Friston2013, Kuchling2020, ramstead2021neural}. We argue that the design of intelligent systems must begin from the \textit{physicality of information and its processing} at every scale or level of self-organization. The result is AI that ``scales up'' the way nature does: by aggregating individual intelligences and their locally contextualized knowledge bases, within and across ecosystems, into ``nested intelligences''---rather than by merely adding more data, parameters, or layers to a machine learning architecture.

We consider the question of how to engineer ecosystems of AI using active inference, with a focus on the problem of communication between intelligent agents, such that shared forms of intelligence emerge, in a nested fashion, from these interactions. We highlight the importance of shared narratives and goals in the emergence of collective behavior, and how active inference helps account for this in terms of sharing (aspects of) the same generative model. We close our discussion with a sketch of stages of development for AI using the principles of active inference. Our hypothesis is that taking the \textit{multi-scale} and \textit{multi-level} aspects of intelligence seriously has the potential to be transformative with respect to the assumptions and goals of research, development, and design in the field of AI, with potentially broad implications for industry and society: that technologies based on the principles described in this paper may be apt to foster the design of an emergent ecosystem of intelligences spanning spatial and cognitive domains (a hyper-spatial web).

\section{A first-principles approach to multi-scale artificial intelligence}\label{sec:first_principles}

The field of artificial intelligence has from the outset used natural systems, whose stunning designs have been refined over evolutionary timescales, as templates for its models. Neuroscience has been the most significant source of inspiration, from the McCulloch-Pitts neuron \cite{Maass1997} to the parallel distributed architectures of connectionism and deep learning \cite{Lecun2015, Lillicrap2019}, to the contemporary call for ``Neuro-AI'' as a paradigm for research in AI, in particular machine learning \cite{Zador2022}. Indeed, the definitive aspect of deep learning inherits from the hierarchical depth of cortical architectures in the brain \cite{Zeki1988}. More recently, machine learning has come, in turn, to influence neuroscience \cite{Yamins2014, Yamins2016UsingGD, Richards2019}. 

Academic research as well as popular media often depict both AGI and ASI as singular and monolithic AI systems, akin to super-intelligent, human individuals. However, intelligence is ubiquitous in natural systems---and generally looks very different from this. Physically complex, expressive systems, such as human beings, are uniquely capable of feats like explicit symbolic communication or mathematical reasoning. But these paradigmatic manifestations of intelligence exist along with, and emerge from, many simpler forms of intelligence found throughout the animal kingdom, as well as less overt forms of intelligence that pervade nature. 

Examples of ``basal cognition'' abound---and often involve distributed, collective forms of intelligence. Colonies of slime molds, for example, can---as a group---navigate two-dimensional spatial landscapes, and even solve mathematical problems that are analytically intractable \cite{Murugan2021}. Certain forms of cognition and learning are (at least arguably) observable in plants \cite{Calvo2017}, and we know that plants grow in a modular fashion, as a structured community of tissues that self-organize into a specific configuration \cite{leyser2011auxin}. Communication between organisms is often mediated by network structures, which themselves consist of other organisms; for instance, it is known that mycorrhizal networks are able to facilitate communication, learning, and memory in trees \cite{Simard2018}. Mobile groups of schooling fish can, as a collective, sense light gradients over a wide spacetime window, even as the individuals that comprise the group can only detect local light intensity \cite{Berdahl2013}. Perhaps most germanely, in morphogenesis (i.e., pattern formation in multicellular organisms), the communication of collectives of cells implements a search for viable solutions in a vast problem space of body configurations \cite{Fields2022d, Davies2022, Fields2020}. This is not merely a metaphorical extension or use of the word ``intelligence,'' as it is no different, at its core, from our experience of navigating three-dimensional space \cite{Kuchling2020}. 

Thus, at each physical spatiotemporal scale of interest, one can identify systems that are competent in their domain of specialization, lending intelligence in physical systems a fundamentally multi-scale character \cite{Levin2019, Fields2021}. Observation of nature suggests, moreover, that simpler and more complex forms of intelligence are almost always related compositionally: appreciably intelligent things tend to be composed of systems that are also intelligent to some degree. Most obviously, the intelligence of individual human beings, to the extent that it depends on the brain, implements the collective intelligence of neurons---harnessed by many intervening levels of organization or modularity, and subserved by organelles at the cellular level. This communal or collective aspect of intelligence is reflected in the etymology of ``intelligence''---from \textit{inter-} (which means between) and \textit{legere} (which means to choose or to read)---literally inter-legibility, or the ability to understand one another.

Since intelligence at each scale supervenes on, or emerges from, simpler (though still intelligent) parts, the multi-scale view of natural intelligence implies not a mysterious infinite regress, but a recursive, nested structure in which the same functional motif (the action-perception loop) recurs in increasingly ramified forms in more complex agents  \cite{Functionalism2008-FUNWG}. The emergence of a higher-level intelligence---from the interaction of intelligent components---depends on \textit{network structure} (e.g., the organization of the nervous system, or communication among members in a group or population) and \textit{sparse coupling} (i.e., the fact that things are defined by what they are \textit{not} connected to \cite{sakthivadivel2022weak}), which together often lead to functional specialization among the constituents \cite{Parr2020b}. 

But how do we engineer systems like these? We argue that instead of focusing merely on empirical descriptive adequacy and ``reasoning by analogy'' (e.g., the Turing test or imitation game \cite{Harnad2003}), one can leverage the fundamental organizing principles that underwrite the operation of intelligence in nature, separating them from the contingent details of particular biological systems. Such an approach has its origins in the cybernetics movement of the 1940s, which set out to describe the general properties of regulatory and purposive systems---that is, properties not tied to any given specific architecture---and from which we draw now commonplace principles of system design, such as feedback and homeostasis \cite{Rosenblueth1943, Ashby1956}. Perhaps the most well-known example of this is the good regulator theorem \cite{Conant1970}, later developed as the internal model principle in control theory \cite{Francis1976}, according to which systems that exist physically must contain structures that are homomorphic to whatever environmental factors they are capable of controlling. A precursor to the good regular theorem is Ashby's ``law of requisite variety'' \cite{Ashby1958}, according to which a system that controls an environment (as represented by a set of observations) must possess at least as many degrees of freedom (in probabilistic terms, as much entropy) as the phenomenon controlled.

Contemporary developments in the statistical mechanics of far-from-equilibrium systems (and in particular, multi-scale, living systems) allow us to formalize these insights---of early cybernetics---as a physics of self-organization, which enables the study of intelligence itself as \textit{a basic, ubiquitous, physical phenomenon}.\footnote{Researchers in AI have often borrowed tools from physics, such as Hamiltonian mechanics, to finesse the inference problems that they face, leading to tools like the Hamiltonian Monte Carlo algorithm, which massively speeds up certain kinds of inferential problem-solving \cite{neal2011mcmc}. Conversely, AI has been used in physics, chemistry, and biochemistry to great effect, allowing us to simulate the containment of plasma in Tomahawk nuclear fusion reactors \cite{Degrave2022}, or predict the ways in which proteins will fold, as the famous AlphaFold system enables \cite{Jumper2021}. What we have in mind, however, is not to borrow techniques or formalisms from physics to solve the problem of intelligent systems design, or to use AI to help finesse problems from physics; but rather, in a complementary fashion, to treat the study of intelligence itself as a chapter of physics.} This has been called a \textit{physics of sentient systems}; where ``sentient'' means ``responsive to sensory impressions'' \cite{Friston2019, Ramstead2019}. More specifically, we argue that one can articulate the principles and corollaries of the core observation that intelligent systems (i.e., agents) exist in terms of a ``Bayesian mechanics'' that can be used to describe or simulate them \cite{DaCosta2021, Ramstead2022a}.

We note that physical implementation is the ultimate constraint on all forms of engineered intelligence. While this claim might sound trivial, it has been a core locus of recent progress in our understanding of the physics of information itself. According to Landauer’s principle \cite{Landauer1961, Bennett2003, Jarzynski1997, Evans2003}, there is an energy cost to irreversibly read-write any information in a physical medium. Thus, the physicality of information and its processing at every scale of self-organization should be accounted for in the design of intelligent systems. Apart from being principled, forcing models to respect constraints or conservation laws---of the kind furnished by physical implementation---often improves their performance or even enables unique capabilities.\footnote{Simulated neural networks, for example, often overfit and fail to generalize if they are not forced to learn compressed representations of their inputs \cite{Schmidhuber2010, Wallace1999, Mackay1995, Olshausen1996}. Relatedly, ubiquitous forms of regularization can be motivated from physical considerations about the finite bandwidth of neurons \cite{Sengupta2010}, and schemes such as predictive coding and sparse coding by considerations about efficient signal transmission \cite{Olshausen1996, Elias1955, Rao1999, Optican1987, Barlow1961, Simoncelli2001}.} Our core thesis is that all of this is naturally accommodated by an approach to AI grounded in the physics of intelligence.

\section{Active inference}\label{sec:actinf}

\subsection{``Model evidence is all you need''}\label{sec:intelligence_inference}

We approach the challenges just outlined from the perspective of active inference, a first-principles or physics-based approach to intelligence that aims to describe, study, and design intelligent agents \textit{from their own perspective} \cite{Eliasmith2005}. Active inference shares the same foundations as quantum, classical, and statistical mechanics, and derives a scale-free theory of intelligence by adding an account of the individuation of particular things within their environments \cite{Friston2019}.

We begin with the observation that individual physical objects can be defined by the typical \textit{absence of influence} of some parts of the universe on others (for example, air temperature directly impacts my skin, but not my internal organs). In \textit{sparse} causal networks, some nodes act as informational bottlenecks that serve both as mediating channels and as (probabilistic) boundaries \cite{Holland2014}, on which the variability of states on either side is conditioned. The persistence of such stable boundaries in a changing world (i.e., away from thermodynamic equilibrium) is possible only to the extent that the boundary conditions can be predicted and controlled, leveraging an implicit statistical model---a \textit{generative model} of how they are caused by external changes.\footnote{In the context of scientific modeling, a statistical model is a mathematical object that encodes the way that things change, relative to the way that other things change. Formally, the structure that encodes such contingencies is called a joint probability distribution. This is the generative model.} 

To exist as an individuated thing is thus to gather observational evidence for such a model (``self-evidencing'' \cite{Hohwy2016}). This ``model evidence'' can be scored by a scalar value that conveys the degree to which some observations conform to (i.e., are predictable from) the model. To account for perception, one can update variables in order to maximize model evidence (e.g., update beliefs to match the data). To account for learning, one can update parameters in order to maximize model evidence (e.g., update models to match the data). To account for action, one can select actions in order to maximize (expected) model evidence (assuming that the model encodes preferences in terms of prior beliefs) \cite{friston2011embodied, Ramstead2019}. From this perspective, model evidence is the only thing that needs to be optimized.

Importantly, model evidence can be approximated in a form that has interesting decompositions, into quantities that map onto distinct facets of intelligence. For instance, a lower bound on the model evidence (equivalently, an upper bound on surprise)---called \textit{variational free energy} \cite{Winn2005}---can always be written as complexity minus accuracy. When a system minimizes free energy, in so doing, it automatically maximizes the predictive accuracy of its model while minimizing its complexity (implementing a version of Occam’s razor). This means that self-evidencing mandates an accurate account of sensory exchanges with the world that is minimally complex, which serves to limit overfitting and poor generalization \cite{Sengupta2018}.

Active inference builds on these insights. If inference entails maximizing accuracy while minimizing complexity, it follows that self-evidencing should minimize the inaccuracy and complexity that is expected following upon a course of action. It transpires that expected complexity is exactly the same quantity minimized in optimal control theory \cite{Kappen2012, Todorov2008}; namely, \textit{risk}, while expected inaccuracy is just the \textit{ambiguity} inherent in the way we sample data (e.g., resolved by switching the lights on in a dark room). Perhaps more interestingly, the ensuing \textit{expected free energy} can be rearranged into expected information gain and expected value, where value is just the (log) preference for an outcome. This result  captures exactly the dual aspects of Bayes optimality; namely, optimal Bayesian experimental design \cite{Mackay1992, Lindley1956, Balietti2021} and decision theory \cite{Berger1985}. In essence, it favors choices that ensure the greatest resolution of uncertainty, under the constraint that preferred outcomes are realized. In other words, it mandates information and preference-seeking behavior, where one contextualizes the other. The ensuing curiosity or novelty-seeking thus emerges as an existential imperative \cite{Mackay1992, Lindley1956, Schwartenbeck2019, Schmidhuber2006, Still2012, Barto2013}---to the extent that one could say that to be intelligent is (in part) to be curious, and to balance curiosity against preferences or reward in an optimal fashion.

Crucially, the approach to existence as modeling just outlined can be applied recursively, in a nested fashion, to systems as well as their components, providing the foundations for mathematical theories of collective intelligence at any scale, from rocks to rockstars.\footnote{Even rocks, while not agents \textit{per se}, track the state of their environment: for instance the interior of a rock ``knows'' that the environment must be well below the melting point of rock (albeit not under that English description). As systems become more elaborate, they can represent more about the things to which they couple \cite{Dennett1983}.} Indeed, if existing in a characteristic way just is soliciting or generating evidence for our existence, then everything that exists can be described as engaging in inference, underwritten by a generative model. Dynamics quite generally can then be cast as a kind of belief updating in light of new information: i.e., changing your mind to accommodate new observations, under the constraint of minimal complexity.

\subsection{AI designed for belief updating}\label{sec:ai_belief_updating}

The principles of natural design that we've reviewed suggest that next-generation AI systems must be equipped with \textit{explicit beliefs} about the state of the world; i.e., they should be designed to implement, or embody, a specific perspective---a perspective under a generative model entailed by their structure (e.g., phenotypic hardware) and dynamics. (Later, we will suggest that efforts should also be directed towards research and development of communication languages and protocols supporting ecosystems of AI.)

\begin{figure}[t!]
    \centering
    \includegraphics[width=1.0\textwidth]{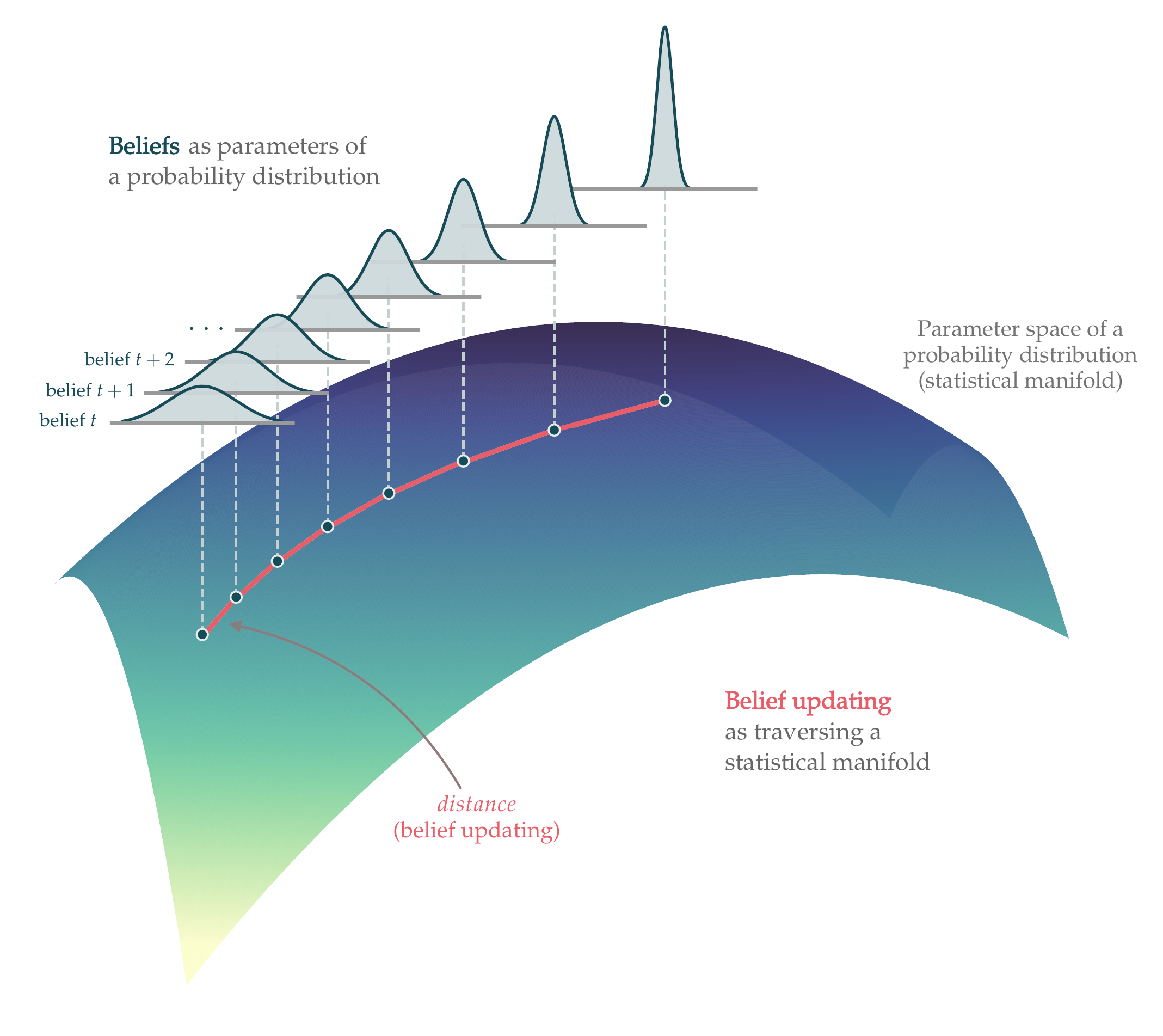}
    \caption{\textbf{Belief updating on a statistical manifold.}}
    \label{fig:info_geo}
\end{figure}

A formal theory of intelligence requires a calculus or mechanics for movement in this space of beliefs, which active inference furnishes in the form of Bayesian mechanics \cite{Ramstead2022a}. Mathematically, belief updating can be expressed as movement in an abstract space---known as a statistical manifold---on which every point corresponds to a probability distribution \cite{Crooks2007, Kim2018, Ay2015, Amari1998, Caticha2015, Parr2020a}. See Figure \ref{fig:info_geo}. This places constraints on the nature of \textit{message passing} in any physical or biophysical realization of an AI system \cite{Kschischang2001, Winn2005, Dauwels2007, Friston2018deep, Parr2019}: messages must be the sufficient statistics or parameters of probability distributions (i.e., Bayesian beliefs). By construction, these include measures of uncertainty. Any variable drawn from a distribution (e.g., the beliefs held by agents about themselves, their environment, and their possible courses of action) are associated with a measure of confidence, known as precision or inverse variance. Thus, intelligent artifacts built according to these principles will appear to quantify their uncertainty and act to resolve that uncertainty (as in the deployment of attention in predictive coding schemes \cite{Feldman2010, Hohwy2012, Kok2011, Kanai2015, Limanowski2022}). Uncertainty quantification is particularly important when assessing the evidence for various models of data, via a process known as structure learning or Bayesian model comparison \cite{Gershman2010, Tenenbaum2011, Spiegelhalter2002, Penny2012, Friston2018}.

There are several types of uncertainty at play when learning from data. First, there may be irreducible noise in the measurement process itself. Examples of such noise include pixel blur in images. Second, the values of the hidden variables being estimated from data may be ambiguous (e.g., ``Is the image I'm viewing of a duck or a rabbit?'' or ``It looks like rain: should I bring an umbrella?''). Third, there may be noise in the model of the function being learned (e.g., ``What do rabbits look like? How do hidden variables map to data?''). Overcoming and accounting for these different types of uncertainty is essential for learning. 

Non-probabilistic approaches to AI encounter these forms of uncertainty but do not represent them explicitly in the structure or parameters of their functions. These methods thus hope to learn successfully without quantifying uncertainty, which is variably feasibile depending on the specific data and output being learned. AI systems that are not purpose-built to select actions in order to reduce uncertainty in an optimal manner will struggle to assign confidence to their predictions. Further, as users of these kinds of AI systems, \textit{we} have no way of knowing how confident they are in their assignments of probability---they are ``black boxes''. 

Taken together, the probabilistic approach provides a normative theory for learning---starting from the first principles of how AI should deal with data and uncertainty. The downside to probabilistic modeling is that it induces severe computational challenges. Specifically, such models must marginalize all the variables in the model in order to arrive at exact ``beliefs'' about a given variable. Thus, the main computational task in probabilistic inference is marginalization, whereas in traditional AI it is the optimization of parameters. As such, a focus on optimization \textit{per se} in contemporary AI research and development may be misplaced to some extent. Current state-of-the-art AI systems are essentially general-purpose optimization machines, built to handle a specific task domain. But optimization in and of itself is not the same as intelligence. Rather, in an intelligent artifact, optimization should be a \textit{method in the service of optimizing our beliefs} about what is causing our data. Fortunately, there are mathematical tricks, such as variational inference, which convert the (intractable) problem of marginalization into a (tractable) optimization problem, allowing probabilistic approaches to utilize the wealth of techniques available for optimization while retaining the benefits of uncertainty quantification. 

\subsection{Comparison to current state-of-the-art approaches}\label{sec:comparison}

Active inference is a very general formulation of intelligence, understood as a self-organizing process of inference. Yet the generality of the formulation is integrative, rather than adversarial or exclusive: it formally relates or connects state-of-the-art approaches (e.g., it has been shown that all canonical neural networks minimize their free energy; \cite{isomura2022canonical}), showcasing their strengths, enabling cross-pollination, and motivating refinements. 

\subsubsection{Managing complexity}\label{sec:ai_complexity}

In the context of machine learning, the complexity term derivable from model evidence (a.k.a., information gain) is especially interesting \cite{Spiegelhalter2002}, since it means that active inference puts predictive accuracy and complexity on an equal footing. In brief, self-evidencing bakes complexity into the optimization of beliefs about the world in a way that automatically finesses many problems with machine learning schemes that focus solely on accuracy \cite{Hochreiter1997b}. To take a salient example from recent discussions, many of the considerations that seem to motivate non-generative approaches---to learning world models \cite{Lecun2022Path}---stem from considering only the likelihood in generative models, rather than model evidence or marginal likelihood---whereas the inclusion of complexity encourages a model to find parsimonious explanations of observations, abstracting from useless detail. In other words, accuracy comes at a (complexity) cost, which must be discounted.

Complexity minimization also speaks to the importance of dimensionality reduction and coarse-graining as clear ways to learn the structure of generative models \cite{Gershman2010}. This is motivated by the intuition that, while sensor data itself is extremely high-dimensional, noisy, unpredictable, ambiguous, and redundant, there is a description of the data in terms of its generating causes (e.g., the world of objects with defined properties) that is lower-dimensional, more predictable, and less ambiguous. Such a description furnishes a compressed, and therefore more efficient, account of the data at hand. Thus, while scaling up data spaces, one may have to scale down the number of latent states generating those data, to the extent that doing so does not sacrifice accuracy. 

Relatedly, active inference provides a promising framework for learning representations in which distinct generative factors are \emph{disentangled} \cite{DBLP:journals/corr/abs-1812-02230}, via the sensorimotor contingencies associated with controllable latent factors \cite{Tschantz2019LearningAM, DBLP:journals/corr/abs-2108-11762, Hinton2011TransformingA}. Low-dimensional  disentangled representations, in addition to being useful for an AI system itself in achieving its own ends, are more explainable and human-interpretable than generic latent representations.

Finally, by encouraging only as much complexity or uncertainty in the model as is needed to account for the source entropy (i.e. the entropy of the generative distribution over states) \cite{Milleretal2022}, the variational free energy objective satisfies Ashby's law of requisite variety, while also ensuring that no \emph{more} complexity exists in a system than is needed to accurately predict observations. Moreover, the need for efficient factorizations of the variational density favor hierarchical, multi-scale systems of the kind we have been describing.

In such systems, patterns or ensembles at the super-ordinate level will possess a certain degree of complexity (as measured by available degrees of freedom or generative factors) that is requisite to model observations at that scale. This entails variability in the internal states of a system, sufficient to model relevant sources of external variability (this can also be motivated as a version of the principle of (constrained) maximum entropy \cite{PhysRev.106.620, Sakthivadivel2022g-a}: 
the optimal `beliefs'---encoded by internal states---are maximally entropic, given constraints supplied by the generative model). But the internal states at one scale just are what individuals at the lower scale are trying to model---so we should expect diversity among the agents participating in any such collective. Simultaneously, sociality, as implemented via belief-sharing (which is necessary to the degree that we are to be predictable to one another) limits variety or entropy, and amounts to an accuracy constraint. In such a system, the sharing of beliefs broadens the evidence base available to each agent (I learn as much about the world by listening to others as by direct observation), but with built-in constraints on both conformity and eccentricity of belief (radical or unusual beliefs may emerge, but they cannot by definition be the norm in the population)---as agents both ``infer together'' (as part of a larger model) and ``infer each other'' (as each constitutes part of the external environment for the others) \cite{parr2020choosing}. 

Minimizing complexity (i.e., compression) thus points to a direction of travel to the Holy Grail of a generic and robust AI; a move from ``big data'' to ``smart data'' or frugal data sampling, based on the principles of multiscale biological intelligence. This has important implications for hardware and energy efficiency. Because the complexity cost has an accompanying thermodynamic cost---via Landauer’s principle and the Jarzynski equality \cite{Jarzynski1997}---there is a lower bound on the thermodynamic cost of any belief updating that can, even in principle, be realized with the right updating scheme. Using active inference, belief updating can be implemented with biomimetic efficiency, without the need for traditional, GPU-based high-performance computing and accompanying costs.

\subsubsection{Reinforcement learning and active inference}\label{sec:ai_rl}

State-of-the-art AI designed for action selection typically implements \textit{reinforcement learning}, a set of methods for maximizing the expected sum of rewards under a sequence of actions. From a Bayesian perspective, however, curiosity and exploration are as fundamental to intelligence as maximizing reward. Specifically, the epistemic, exploratory, curiosity-driven aspect of intelligence motivates actions expected to reduce uncertainty in the variables and parameters that define your model; which, in the active inference formulation, corresponds to inference and learning, respectively \cite{Mackay1992, Lindley1956, Schwartenbeck2019, Schmidhuber2006, Still2012, Barto2013}. 

In line with the above discussion of self-evidencing, rather than select actions that maximize some arbitrary state-dependent reward (or equivalently, minimize an arbitrary cost function), an intelligent system ought to generate observations or sensory data consistent with its characteristic (i.e., preferred) exchanges with the sensed world, and thus with its own continued existence. That is, an intelligent agent ought to maximize the evidence for its generative model. Active inference thereby generalizes the notion of reward, and labels every encountered outcome (and implicitly every latent state) in terms of how likely it is that ``this would happen to me''. This is scored in terms of prior preferences over outcomes, which are part of the generative model. Preferences over some kinds of outcomes are precise (e.g., not being eaten or embarrassed), others less so (``I prefer coffee in the morning, but tea is nice''). To summarize, preferences provide constraints that define the ``kind of thing I am,'' with more precise preferences playing a similar role, for example, to the ``intrinsic costs'' in the actor-critic system proposed in \cite{Lecun2022Path}.

In this view, Bayesian reinforcement learning is a \textit{special case} of active inference, in which the preference for all outcomes is very imprecise---apart from one privileged outcome called reward, for which there is a very precise preference. The perspective from active inference moves our notion of intelligence away from a monothematic reward optimization problem, towards a multiple-constraint-satisfaction problem, where the implicit ‘satisficing’ \cite{Gigerenzer2010} just is self-evidencing.

\subsubsection{Multi-scale considerations}\label{sec:ai_multiscale}

Another key difference concerns the multi-scale architecture of active inference. First, active inference commits to a separation of temporal scales, which allows it to finesse key issues in AI research. On the present account, learning is just slow inference, and model selection is just slow learning. All three processes operate in the same basic way, over nested timescales, to maximize model evidence. 

Second, active inference predicts, and provides a formalism for describing, the multi-scale character of intelligence in nature; see also \cite{krakauer2020information}. Although this has generally not been a focus of research in machine learning, work in the field consonant with this perspective includes the complex internal structure of LSTM cells \cite{Hochreiter1997b}, the repetition of the split-transform-merge strategy across scales in the ResNeXt architecture \cite{Xie2016}, capsule networks \cite{Sabour2017}, in which individually complex nodes engage in a form of self-organization, the Thousand Brains theory of the cooperation of cortical columns to produce global representations \cite{Hawkins2017}, or the perspective on restricted Boltzmann machines as products of experts \cite{Hinton2002}. 

Relatedly, beyond fixing certain hyperparameters of system design (such as the general character of the objective function to be optimized), active inference is itself silent on the way in which model evidence maximization is implemented in particular systems. For this reason, it is crucial that work within this framework be informed by, and participate in, ongoing threads of research in machine learning and empirical neuroscience. Predictive coding, for example \cite{Rao1999}, is a way of realizing active inference in brains, and perhaps in similar systems with many internal degrees of freedom and shorter-timescale plasticity. Many other aspects of complex intelligence, including quite essential ones with roots deep in evolutionary history, may depend on details of that history difficult to predict from first principles alone---for example, mechanisms within the hippocampal/entorhinal system known to enable spatial navigation and localization may constitute much more general-purpose high-level design patterns for neural systems \cite{safron2022gslam}.

\subsection{Shared narratives}\label{sec:shared_narratives}

We have noted that intelligence as self-evidencing is inherently perspectival, as it involves actively making sense of and engaging with the world from a specific point of view (i.e., given a set of beliefs). Importantly, if the origins of intelligence indeed lie in the partitioning of the universe into subsystems by probabilistic boundaries, then intelligence never arises singly but always exists \textit{on either side} of such a boundary \cite{Constant2018, Veissiere2020}. The world that one models is almost invariably composed of other intelligent agents that model one in turn. 

This brings us back to the insight that intelligence must, at some level, be distributed over every agent and over every scale at which agents exist. Active inference is naturally a theory of collective intelligence. There are many foundational issues that arise from this take on intelligence; ranging from communication to cultural niche construction: from theory of mind to selfhood \cite{Laland1999, Constant2018, Constant2019, Veissiere2020, Vasil2020}. On the active inference account, shared goals emerge from shared narratives, which are provided by shared generative models \cite{bouizegarene2020narrative}. Furthermore---on the current analysis---certain things should then be \textit{curious about each other}. 

The importance of perspective-taking and implicit shared narratives (i.e., generative models or frames of reference) is highlighted by the recent excitement about \textit{generative AI} \cite{Sequoia2022}, in which generative neural networks demonstrate the ability to reproduce the kinds of pictures, prose, or music that we expose them to. Key to the usage of these systems is a \textit{dyadic interaction} between artificial and natural intelligence, from the training of deep neural networks to the exchange of prompts and generated images with the resulting AI systems, and the subsequent selection and sharing of the most apt ``reproductions'' among generated outputs.\footnote{The importance of fluid exchange between artificial and human intelligence in this paradigm is evinced by the rapidly growing interest in \textit{prompt engineering}, i.e., an increasingly self-aware and theory-driven approach to the role that prompts play in co-creating the outputs of these types of systems \cite{DBLP:journals/corr/abs-2107-13586}, which has recently been extended to the optimization of text prompts by distinct AI agents \cite{Deng2022RLPromptOD}.} In our view, a truly intelligent generative AI would then become curious about us---and want to know what we are likely to select. In short, when AI takes the initiative to ask \textit{us} questions, we will have moved closer to genuine intelligence, as seen through the lens of self-evidencing.

\section{From Babel to binary}\label{sec:babel_binary}

Human intelligence and language have co-evolved, such that they both scaffold, and are scaffolded by, one another \cite{Tomasello2016, Heyes2018}. The core functional role of language is to enable communication and shared understanding: language has been optimized for sharing with other intelligent creatures (as a language that can be easily passed on reaches further generations). Language has thus facilitated the emergence of more complex interactions and shared customs between agents, which has in turn allowed for the emergence of intensive human collaboration at multiple communal scales \cite{henrich2015secret}. Relatedly, language provides a reference for how to ``carve nature at its joints'' (e.g., into objects, properties, and events), facilitating learning about the world and the way it works. Finally, it has allowed humans to build an external store of knowledge far beyond the epistemic capacity of any human individual. Human beings both benefit from---and contribute to---this store of knowledge, which, like language itself, has co-evolved with our intelligence. 

Across cultures, the earliest recorded narratives of our species have emphasized the astounding integrative power of shared communication systems along with their flipside: the discord and disarray wrought by miscommunication and a lack of mutual understanding. This is illustrated potently in the biblical story of the Tower of Babel, which tells of a mighty civilization that attempted to build a glorious city with a tower that rose to the heavens. These lofty aspirations fell to ruin after a divine disruption that eliminated their common language, shattering it into a thousand uninterpretable dialects. In their confusion and miscomprehension, they were unable to complete the Tower and were thus scattered across the Earth, forced to survive in the clustered tribes that shared their regional vernacular. 

Today, humans cope with a ``post-Babel'' world via a combination of increasing multilingualism, rallying (for better or worse) behind hegemonic languages like English, and, recently, increasingly effective machine translation \cite{DBLP:journals/corr/WuSCLNMKCGMKSJL16}. Digital computers \textit{do} share a common or universal machine language (i.e., binary representation). If situations can be represented adequately in an appropriate machine syntax, they can be subjected to the operations of mathematical logic, formalized and thereby processed in an unambiguous way. At a higher level, it may be said that ``vectorese'' is the universal language of AI, in that vectors (i.e., ordered lists of numbers representing a point in an abstract space) constitute the input, output, and medium of data storage and retrieval for most AI algorithms. 

Vectors are analogous to the medium of action potentials in the brain---they are capable of representing anything we can think of, but nearly all the interesting (and representationally load-bearing) structure lies in the (learned) transformations and accompanying transition dynamics of the underlying dynamical system. Often, an output vector space can be considered as an embedding or transformation of the input space, and mappings among vector spaces are much like translations among languages. However, vectors themselves may only provide a base structure or medium (analogous to sound or light) for higher-level languages. 

It has been clear from the early days of neural language modeling that vector space representations can in principle be learned that capture both the semantic and syntactic regularities implicit in the co-occurrence statistics of natural language corpora \cite{DBLP:journals/corr/abs-1103-0398, DBLP:journals/corr/MikolovSCCD13}. Despite this, we lack anything like \textit{a common high-level language} that AIs can use to communicate with one another and with humans---other than, arguably, human natural languages themselves, which can be used to interface with AIs via modern language models. The fact that reinforcement learning agents trained to produce prompts for such models often produce unintelligible nonsense strings, however \cite{Deng2022RLPromptOD, Webson2022DoPM}, shows that even where large language models use English, they do not use or understand it in the way humans do; this raises the question whether natural languages can really play the role of a shared human-machine language without modification.

Moreover, while the necessity of serializing thought into discrete token sequences for the purposes of communication helps enforce the kind of sparsity structure that we have argued is essential to intelligence and complexity itself, a more direct form of information transfer is also conceivable in which the richness of a latent vector representation (or ``thought'') is directly externalized as a data structure. While current state-of-the-art AI can learn the language of vector space embeddings, the science of inter-AI communication and shared latent spaces is in its infancy. For the most part, each AI must learn to carve up the world from scratch, and is unable to share its knowledge fluidly or update it in collaboration with other AIs.\footnote{An important exception is the proliferation of fine-tuned copies of large monolithic pre-trained models such as BERT. This is not obviously relevant to our interest in (possibly real-time) communication and mutual updating among persistent, physically situated AI systems, though it may constitute a form of evolution of populations of AI systems with partially divergent learning histories.}

We argue that the future evolution of AI would benefit greatly from a focus on optimization for \textit{shareability} (i.e., gathering evidence for a model of an intrinsically \textit{social} creature.) This might take the form of a shared external store of knowledge about \textit{how to communicate with relevant others}, or a structured communication protocol that can act as the \textit{lingua franca} of AI. A general framework that ties together different embedding spaces and inter-AI messaging over a shared network architecture would, among other things, enable AI agents to learn to offload certain tasks or predictions to other, more specialized AI agents. 

\subsection{Active inference and communication}\label{sec:actinf_communication}

An underlying theme thus far is that intelligence at any scale requires a shared generative model and implicit common ground. There are many ways to articulate this theme; from ensemble learning to mixtures of experts \cite{Hinton2002}, from distributed cognition to Bayesian model averaging \cite{Fitzgerald2014}. 

Imagine that someone has locked you in a large dark room. As a self-evidencing and curious creature, you would be compelled to feel your way around to resolve uncertainty about your situation. Successive palpations lead you to infer that there is a large animal in the room---by virtue of feeling what seem to be a tail, a succession of legs, and eventually a trunk. Your actions generate accumulated evidence for the hypothesis ``I am in a room with an elephant.'' Now, imagine an alternative scenario in which you and five friends are deployed around the same room, and can report what you feel to each other. In this scenario, you quickly reach the consensus ``We are in a room with an elephant.'' The mechanics of belief updating are similar in both scenarios. In the first, you accumulate evidence and successively update your posterior belief about latent states. In the second, the collective assimilation of evidence is parallelized across multiple individuals. 

Is the latter equivalent to having one brain with twelve hands? Not quite. The second kind of belief updating rests upon a shared generative model or hypothesis space that enables you to assimilate the beliefs of another. For example, you share a common notion of a ``trunk,'' a ``leg,'' and a ``tail''---and crucially, you have access to a shared language for communicating such concepts. Sharing a generative model allows each agent to infer the causes of its sensations and disentangle the causes that are unique to the way the world is sampled---e.g., ``where I am looking''---and causes that constitute the shared environment (e.g., ``what I am looking at'') \cite{Veissiere2020, Ramstead2016, Friston2015a}. Just as importantly, any dyad or ensemble of self-evidencing agents will come to share a generative model (or at least some factors of a generative model) via their interactions \cite{sakthivadivel2022weak} (see \cite{Friston2015b, Isomura2019} for numerical experiments in active inference that illustrate this phenomenon, and Table \ref{tab:actinf_applications} for related applications.) 

What results is a shared intelligence (i.e., a kind of collective super-intelligence) that emerges from an ensemble of agents. Heuristically, maximizing model evidence means making the world as predictable as possible \cite{Albarracin2022, Kastel2021}. This is assured if we are both singing from the same hymn sheet, so to speak---so that I can predict you and you can predict me. Mathematically, this is evinced as a generalized synchrony between the dynamics on our respective statistical manifolds \cite{Friston2015b,Palacios2019}. This generalized synchrony (or synchronicity) is special because it unfolds in a (shared) belief space, meaning it can be read as mutual understanding: i.e., coming to align our beliefs, via a shared language and a shared generative model. This sharedness is arguably the basis of culture and underpins the existence of our civilization. Our challenge, which we take to be a necessary step toward ASI or even AGI, is to expand the sphere of culture to include artificial agents.

\subsection{Belief propagation, graphs, and networks}\label{sec:belief_graph_networks}

Operationally, ecosystems of shared intelligence can be described in terms of message passing on a factor graph \cite{Winn2005, Dauwels2007, Friston2017a, Yedidia2005}, a special kind of graph or network in which nodes correspond to the factors of a Bayesian belief or probability distribution. Factors are just probabilistic beliefs that one multiplies together to get a joint distribution (i.e., a generative model). For example, one could factorize beliefs about the latent states of an object into ``what'' and ``where.'' These beliefs jointly specify a unique object in extrapersonal space; noting that knowing what something is and knowing where it is are largely independent of each other \cite{Ungerleider1994}. The edges of a factor graph correspond to the messages passed among factors that underwrite belief updating. In the implementations of active inference that we have been describing, they comprise the requisite sufficient statistics that summarize the beliefs of other nodes. 

Technically, this is useful because for any generative model there is a dual or complementary factor graph that prescribes precisely the requisite message passing and implicit computational architecture. In our setting, this architecture has an interesting aspect: we can imagine the nodes of a vast graph partitioned into lots of little subgraphs. Each of these would correspond to an agent updating its beliefs via the propagation of internal messages. Conversely, external messages would correspond to communication and belief-sharing that rests upon certain factors being distributed or duplicated over two or more subgraphs (i.e., agents or computers). This kind of architecture means that, in principle, any subgraph or agent can see, vicariously, every observable in the world---as seen through the eyes of another agent. But what is the functional and structural form of the generative model that underwrites such an architecture?

Taking our lead from human communication, the most efficient (minimum description length or minimum-complexity) generative model of worldly states should be somewhat simplified (i.e., coarse-grained), leveraging discrete representations with only as much granularity as is required to maintain an accurate account of observations \cite{Heins2022spin, Klein2020emergence}. There are many motivations for this kind of generative model. First, it is continuous with the approach to thing-ness or individuation described above, according to which individuals are defined by the sparsity of their interactions. Concepts should evince a sparse structure, both because they are themselves ``things'' (and so should have sparse connections to other similar ``things''), and because they are accurate representations of a world characterized by sparsity. Second, belief updating can, in this case, be implemented with simple linear operators, of the sort found in quantum computation \cite{Fields2021, Fields2022c, Parrondo2015}. Furthermore, this kind of discretization via coarse-graining moves us into the world of the theory of signs or semiotics \cite{Roy2005, Sewell1992}, Boolean logic and operators, and the sort of inference associated with abductive reasoning \cite{peirce1931collected}. Finally, it finesses the form of message passing, since the sufficient statistics of discrete distributions can be reduced to a list of the relative probabilities of being in the states or levels of any given factor \cite{Ghahramani1997}, enabling AI systems to flexibly switch contexts and acquire knowledge from others quickly and adaptively, based on a repository of shared representations.

\subsection{Intelligence at scale}\label{sec:intelligence_scale}

A subtle aspect of active inference, in this setting, is the selection of which messages to listen or attend to. In principle, this is a solved problem---in the simple case, each agent (i.e., subgraph) actively selects the messages or viewpoints that afford the greatest expected information gain \cite{Friston2020a}.\footnote{See \cite{Albarracin2022} for more complex cases where agents have preferences for certain kinds of interaction partners, resulting in the formation of ``echo chambers.''} The neurobiological homologue of this would be attention: selecting the newsworthy information that resolves uncertainty about things you do not already know, given a certain context. There are many interesting aspects of this enactive (action-oriented) aspect of message passing; especially when thinking about nested, hierarchical structures in a global (factor) graph. In these structures---and in simulations of hierarchical processing in the brain---certain factors at higher hierarchical levels can control the selection of messages by lower levels \cite{Parr2017, Smith2019}. This motivates exploration of the multi-scale aspects of shared intelligence. 

The emerging picture is of a message passing protocol that instantiates variational message passing on graphs of discrete belief spaces. But what must these messages contain? Clearly, on the present proposal, they must contain vectors of sufficient statistics; but they also have to identify themselves in relation to the (shared) factors to which they pertain. Furthermore, they must also declare their origin, in much the same way as neuronal populations in the brain receive spatially addressed inputs from other parts of the brain. 

In a synthetic setting, this calls for spatial addressing, leading to the notion of a \textit{spatial message passing protocol and modeling language}---of the sort being developed as open standards in the Institute of Electrical and Electronics Engineers (IEEE) P2874 Spatial Web Working Group \cite{SpatialWebWorkingGroup}. In short, the first step---toward realizing the kind of distributed, emergent, shared intelligence we have in mind---is to construct the next generation of modeling and message passing protocols, which include an irreducible spatial addressing system amenable to vectorization, and allowing for the vector-based shared representation of much of human knowledge.

\section{Ethical and moral considerations}\label{sec:ethics}

We conclude our discussion of large-scale collective intelligence with a brief discussion of the relevant areas of ethical discussion---and contention. First, it is important to note that the kind of collective intelligence evinced by eusocial insects (e.g., ant colonies), in which most individuals are merely replaceable copies of one another, is not the only paradigm for shared intelligence---nor is it a suitable one for systems in which individual nodes embody complex generative models. We believe that developing a cyber-physical network of emergent intelligence in the manner described above not only ought to, but for architectural reasons \textit{must}, be pursued in a way that positively values and safeguards the \textit{individuality} of people (as well as potentially non-human persons). 

This idea is not new. Already in the late 1990s, before the widespread adoption of the internet as a communication technology, a future state of society had been hypothesized in which the intrinsic value of individuals is acknowledged in part because knowledge is valuable and \textit{knowledge and life are inseparable} \cite{Levy1997}---that is, each person has a distinct and unique life experience and, as such, knows something that no one else does. This resonates deeply with our idea that every intelligence implements a generative model of its own existence. The form of collective intelligence that we envision can emerge only from a network of essentially unique, epistemically and experientially diverse agents. This useful diversity of perspectives is a special case of functional specialization across the components of a complex system.

Much discussion in the field of AI ethics focuses on the problem of AI alignment; i.e., aligning our value systems with those of hypothetical conscious AI agents, which may possibly evince forms of super-intelligence \cite{russell2019human, allen2005artificial, russell2015letter}; for critical discussion, see \cite{marcus2019rebooting}. This can be discussed under the broader rubric of the capacity for empathy or sympathy---what one might call \textit{sympathetic} intelligence---which concerns the ability of agents to share aspects of their generative models, to take the perspective of other agents, and to understand the world in ways similar enough to enable coordinated action. This likely requires avoiding undesirable equilibria, such as those evincing pathologies of alignment (e.g., the elimination of a healthy diversity of perspectives), as well as those resembling the predator-prey systems found in nature\footnote{We thank George Percivall for raising these points.}. Whether the emergence of shared intelligence in such a network structure entails the emergence of a new, collective mind is an open question.

Current state-of-the-art AI systems are largely ``black boxes.'' Such an approach to the design of AI ultimately puts severe limits on its transparency, explainability, and auditability. In addition, their capacity to engage in genuine collaboration with humans and other AI is limited, because they lack the ability to take the perspective of another. Moving to multi-scale active inference offers a number of technical advantages that may help address these problems. One is that leveraging explicit generative models, which carve the world into discrete latent states, may help us to identify and quantify bias in our models. Such architectures feature increased auditability, in that they are explicitly queryable and their inferences can be examined forensically---allowing us to address these biases directly. Shared generative models also effectively equip AI with a theory of mind, facilitating perspective-taking and allowing for genuinely dyadic interactions. 

Much like a brain, with its many layers and connections, the multi-scale architecture for collective intelligence that we propose could be equipped with nodes and layers to enable a kind of collective self-monitoring and self-organisation of salience. However, this raises the question of authority and power: this kind of approach to the design of AI must account for the plurality and vulnerability of individual perspectives, and the need to understand and counterbalance potential abuses of power. More broadly and perhaps more fundamentally, we note that the approach to AI that we have presented here does not obviate the dangers associated with bias in AI technologies, especially when deployed at industrial scale in commercial settings, e.g., \cite{birhane2021algorithmic}. The general idea is that the deployment of AI technologies in societies that have preexisting hierarchies of power and authority can have problematic consequences. For example, discriminatory bias encoded in data will result in unfairly biased AI systems (e.g., \cite{birhane2022unseen}) regardless of the specific technologies used to build that AI. It is highly probable that the use of AI technologies premised on such data will sustain social biases and practices that are harmful, or may represent future harm, the consequences of which are not yet fully known---or may be unknowable---regardless of the intentions of the creators. These concerns are well founded and cannot be resolved through narrowly technical means. As such, some combination of novel social policies, government regulations, and ethical norms are likely to be required to ensure that these new technologies harness and reflect our most essential and persistent values.

We are not pessimistic. Nature provides us with endless demonstrations of the success of emergent, shared intelligence across systems at every scale. Looking back to the elegant design of the human body, we find bountiful examples of diverse systems of nested intelligences working together to seek out a dynamic harmony and balance. As an integrated system, the body is capable of achieving multi-scale homeostasis and allostasis, notably via the incredible coordination and communicative power of the nervous system, allowing it to adapt to novel environmental conditions and to regulate its needs in real time. We reiterate our conviction that the design of AI should be informed by, and aligned with, these time-tested methods and design principles. Furthermore, we believe that the class of sympathetic and shared intelligences that we have described in this paper offers a responsible and desirable path to achieving the highest technical and ethical goals for AI, based on a design of ecosystems of intelligence from first principles.

\section{Conclusion: Our proposal for stages of development for active inference as an artificial intelligence technology}\label{sec:conclusion}

The aim of this white paper was to present a vision of research and development in the field of artificial intelligence for the next decade (and beyond). We suggested that AGI and ASI will emerge from the interaction of intelligences networked into a hyper-spatial web or ecosystem of natural and artificial intelligence. We have proposed active inference as a technology uniquely suited to the collaborative design of an ecosystem of natural and synthetic sense-making, in which humans are integral participants---what we call shared intelligence. The Bayesian mechanics of intelligent systems that follows from active inference led us to define intelligence operationally, as the accumulation of evidence for an agent’s generative model of their sensed world---also known as self-evidencing. This self-evidencing can be implemented using message passing or belief propagation on (factor) graphs or networks. Active inference is uniquely suited to this task because it leads to a formal account of collective intelligence. We considered the kinds of communication protocols that must be developed to enable such an ecosystem of intelligences, and argued that such considerations motivate the development of a generalized, \textit{hyper-spatial modeling language and transaction protocol}. We suggest that establishing such common languages and protocols is a key enabling step towards an ecosystem of naturally occurring and artificial intelligences.

In closing, we provide a roadmap for developing intelligent artifacts and message passing schemes as methods or tools for the common good. This roadmap is inspired by the technology readiness levels (TRLs) that have been adopted as a framework for understanding progress in technical research and development by institutions such as the European Commission, the International Organization for Standardization (ISO), and the National Aeronautics and Space Administration agency (NASA).
\subsection{Stages of development for active inference}\label{sec:stages_ai}

\paragraph{S0: Systemic Intelligence.} This is contemporary state-of-the-art AI; namely, universal function approximation---mapping from input or sensory states to outputs or action states---that optimizes some well-defined value function or cost of (systemic) states. Examples include deep learning, Bayesian reinforcement learning, etc.

\paragraph{S1: Sentient Intelligence.} Sentient behavior or active inference based on belief updating and propagation (i.e., optimizing beliefs about states as opposed to states \textit{per se}); where ``sentient'' means ``responsive to sensory impressions.''\footnote{To preempt any worries, we emphasize that we do \textit{not} mean that sentient intelligent systems are necessarily \textit{conscious}, in the sense of having qualitative states of awareness; e.g., as the word was used in the recent controversy surrounding Google's AI system LaMDA \cite{cohen2022lamda}. It is standard to use the word ``sentient'' to mean ``responsive to sensory impressions'' in the literature on the free energy principle; e.g., in \cite{Ramstead2019}}. This entails planning as inference; namely, inferring courses of action that maximize expected information gain and expected value, where value is part of a generative (i.e., world) model; namely, prior preferences. This kind of intelligence is both information-seeking and preference-seeking. It is quintessentially curious.

\paragraph{S2: Sophisticated Intelligence.} Sentient behavior---as defined under S1---in which plans are predicated on the consequences of action for beliefs about states of the world, as opposed to states \textit{per se}. I.e., a move from ``what will happen if I do this?'' to ``what will I believe or know if I do this?'' \cite{Friston2021, Hesp2020}. This kind of inference generally uses generative models with discrete states that ``carve nature at its joints''; namely, inference over coarse-grained representations and ensuing world models. This kind of intelligence is amenable to formulation in terms of modal logic, quantum computation, and category theory. This stage corresponds to ``artificial general intelligence'' in the popular narrative about the progress of AI.

\paragraph{S3: Sympathetic (or Sapient) Intelligence.} The deployment of sophisticated AI to recognize the nature and dispositions of users and other AI and---in consequence---recognize (and instantiate) attentional and dispositional states of self; namely, a kind of minimal selfhood (which entails generative models equipped with the capacity for Theory of Mind). This kind of intelligence is able to take the perspective of its users and interaction partners---it is perspectival, in the robust sense of being able to engage in dyadic and shared perspective-taking. 

\paragraph{S4: Shared (or Super) Intelligence.} The kind of collective that emerges from the coordination of Sympathetic Intelligence (as defined in S3) and their interaction partners or users---which may include naturally occurring intelligence such as ourselves, but also other sapient artifacts. This stage corresponds, roughly speaking, to ``artificial super-intelligence'' in the popular narrative about the progress of AI---with the important distinction that we believe that such intelligence will emerge from dense interactions between agents networked into a hyper-spatial web. We believe that the approach that we have outlined here is the most likely route toward this kind of hypothetical, planetary-scale, distributed super-intelligence \cite{frank2022intelligence}.

\subsection{Implementation}\label{sec:implementation_ai}

\paragraph{A: Theoretical.} The basis of belief updating (i.e., inference and learning) is underwritten by a formal calculus (e.g., Bayesian mechanics), with clear links to the physics of self-organization of open systems far from equilibrium.

\paragraph{B: Proof of principle.} Software instances of the formal (mathematical) scheme, usually on a classical (i.e., von Neumann) architecture.
 
\paragraph{C: Deployment at scale.} Scaled and efficient application of the theoretical principles (i.e., methods) in a real-world setting (e.g., edge-computing, robotics, variational message passing on the web, etc.)
 
\paragraph{D: Biomimetic hardware.} Implementations that elude the von Neumann bottleneck, on biomimetic or neuromorphic architectures. E.g., photonics, soft robotics, and belief propagation: i.e., message passing of the sufficient statistics of (Bayesian) beliefs.

\begin{table}[H]
{\renewcommand{\arraystretch}{1.5}
\resizebox{\columnwidth}{!}{
\begin{tabular}{|l|l|l|l|l|l|}
\rowcolor[HTML]{D7D7D7}
\hline
\textbf{Stage} & \textbf{Theoretical} & \textbf{Proof of principle} & \textbf{Deployment at scale} & \textbf{Biomimetic} & \textbf{Timeframe} \\\hline\hline
\textbf{S1: Sentient} & \cellcolor[HTML]{D4F5D4}Established$^{1,2}$ & \cellcolor[HTML]{D4F5D4}Established$^{3}$ & \cellcolor[HTML]{CBECF5}Provisional$^{4}$ & \cellcolor[HTML]{CBECF5}Provisional & 6 months \\\hline
\textbf{S2: Sophisticated} & \cellcolor[HTML]{D4F5D4}Established$^{5}$ & \cellcolor[HTML]{D4F5D4}Established$^{6}$ & \cellcolor[HTML]{CBECF5}Provisional &  & 1 year \\\hline
\textbf{S3: Sympathetic} & \cellcolor[HTML]{D4F5D4}Established$^{7}$ & \cellcolor[HTML]{CBECF5}Provisional & \cellcolor[HTML]{CBECF5}Provisional &  & 2 years \\\hline
\textbf{S4: Shared} & \cellcolor[HTML]{D4F5D4}Established$^{8,9,10,11}$ & \cellcolor[HTML]{F7E4C0}Aspirational & \cellcolor[HTML]{F7E4C0}Aspirational &  & 8 years\\\hline
\end{tabular}}}
\caption{\textbf{Stages of AI premised on active inference.}}\vspace{0.25cm}
\begin{hangparas}{.35in}{1}
{\footnotesize 
$^1\,$Friston, K.J. A free energy principle for a particular physics. doi:\href{https://doi.org/10.48550/arXiv.1906.10184}{10.48550/arXiv.1906.10184} (2019).\cite{Friston2019}

$^2\,$Ramstead, M.J.D.~et al.~On Bayesian Mechanics: A Physics of and by Beliefs. \textit{Interface Focus} 13, doi:\href{https://doi.org/10.1098/rsfs.2022.0029}{10.1098/rsfs.2022.0029} (2023).\cite{Ramstead2022a}

$^3\,$Parr, T., Pezzulo, G. \& Friston, K.J. Active Inference: The Free Energy Principle in Mind, Brain, and Behavior. (MIT Press, 2022). doi:\href{https://doi.org/10.7551/mitpress/12441.001.0001}{10.7551/mitpress/12441.001.0001}.\cite{Parr2022}

$^4\,$Mazzaglia, P., Verbelen, T., Catal, O. \& Dhoedt, B. The Free Energy Principle for Perception and Action: A Deep Learning Perspective. \textit{Entropy} 24, 301, doi:\href{https://doi.org/10.3390/e24020301}{10.3390/e24020301} (2022).\cite{Mazzaglia2022}




$^5\,$Da Costa, L. et al.~Active inference on discrete state-spaces: A synthesis. \textit{Journal of Mathematical Psychology} 99, 102447, doi:\href{https://doi.org/10.1016/j.jmp.2020.102447}{10.1016/j.jmp.2020.102447} (2020).\cite{DaCosta2020}

$^6\,$Friston, K.J., Parr, T. \& de Vries, B. The graphical brain: Belief propagation and active inference. \textit{Network Neuroscience} 1, 381-414, doi:\href{https://doi.org/10.1162/NETN_a_00018}{10.1162/NETN\_a\_00018} (2017).\cite{Friston2017a}

$^{7}\,$Friston, K.J. et al.~Generative models, linguistic communication and active inference. \textit{Neuroscience and Biobehavioral Reviews} 118, 42-64, doi:\href{https://doi.org/10.1016/j.neubiorev.2020.07.005}{10.1016/j.neubiorev.2020.07.005} (2020).\cite{Friston2020a}

$^{8}\,$Friston, K.J., Levin, M., Sengupta, B. \& Pezzulo, G. Knowing one's place: a free-energy approach to pattern regulation. \textit{Journal of the Royal Society Interface} 12, doi:\href{https://doi.org/10.1098/rsif.2014.1383}{10.1098/rsif.2014.1383} (2015).\cite{Friston2015a}

$^{9}\,$Albarracin, M., Demekas, D., Ramstead, M.J.D. \& Heins, C. Epistemic Communities under Active Inference. \textit{Entropy} 24, doi:\href{https://doi.org/10.3390/e24040476}{10.3390/e24040476} (2022).\cite{Albarracin2022}

$^{10}\,$Kaufmann, R., Gupta, P., \& Taylor, J. An Active Inference Model of Collective Intelligence. \textit{Entropy} 23(7), doi:\href{doi:https://doi.org/10.3390/e23070830}{10.3390/e23070830} (2021).\cite{e23070830}

$^{11}\,$Heins, C., Klein, B., Demekas, D., Aguilera, M., \& Buckley, C. Spin Glass Systems as Collective Active Inference. \textit{International Workshop on Active Inference} 2022, doi:\href{doi:https://doi.org/10.1007/978-3-031-28719}{10.1007/978-3-031-28719} (2022).\cite{Heins2022spin}

}
\end{hangparas}
\label{tab:stages_ai}
\end{table}

\section*{Additional information}\label{sec:additional_info}

\paragraph{Acknowledgements}
The authors thank Rosalyn Moran and George Percivall for useful discussions. Table \ref{tab:actinf_applications} in Appendix \ref{sec:appendix_A} has been reproduced from \cite{DaCosta2020} under a CC BY 4.0 licence (\url{https://creativecommons.org/licenses/by/4.0/}).

\vspace{-0.25cm}
\paragraph{Funding information}
All work on this paper was funded by VERSES. KF is supported by funding for the Wellcome Centre for Human Neuroimaging (Ref: 205103/Z/16/Z) and a Canada-UK Artificial Intelligence Initiative (Ref: ES/T01279X/1). CH is supported by the U.S. Office of Naval Research (Ref: N00014-19-1-2556). BK \& CH acknowledge the support of a grant from the John Templeton Foundation (Ref: 61780). The opinions expressed in this publication are those of the author(s) and do not necessarily reflect the views of the John Templeton Foundation. BM was funded by Rafal Bogacz with a BBSRC grant (Ref: BB/s006338/1) and a MRC grant (Ref: MC UU 00003/1). SET is supported in part by funding from the Social Sciences and Humanities Research Council of Canada (Ref: 767-2020-2276). 

\begin{sloppypar}
\printbibliography[title={References}]
\end{sloppypar}

\clearpage

\appendix
\setcounter{figure}{0}
\setcounter{table}{0}
\setcounter{equation}{0}
\renewcommand\thefigure{\thesection.\arabic{figure}}
\renewcommand\thetable{\thesection.\arabic{table}}
\renewcommand\theequation{\thesection .\arabic{equation}}
\begin{refsection}
\setcounter{page}{1}

\section{Appendix: Applications of active inference}\label{sec:appendix_A}

\small
\begin{center}
{\renewcommand{\arraystretch}{1.5}
\begin{longtable}{|L{0.21\linewidth} | L{0.33\linewidth}| L{0.26\linewidth}|}
\caption{\textbf{Examples of Active Inference implementations.} From Da Costa et al.~(2020) \cite{DaCosta2020}}
\label{tab:actinf_applications}\\
\endfirsthead
\hline
\textbf{Application} & \textbf{Description} & \textbf{References} \\\hline\hline
Decision-making under uncertainty & Initial formulation of active inference on partially observable Markov decision processes. & \href{https://doi.org/10.1007/s00422-012-0512-8}{Friston, Samothrakis et al. (2012)} \cite{Friston2012} \\\hline
Optimal control & Application of KL or risk sensitive control in an engineering benchmark---the mountain car problem. & \href{https://ieeexplore.ieee.org/document/9054364}{Çatal et al.~(2019)} \cite{Catal2020} and \href{https://doi.org/10.3389/fnbot.2012.00011}{Friston, Adams et al.~(2012)} \cite{Friston2012value} \\\hline
Evidence accumulation & Illustrating the role of evidence accumulation in decision-making through an urns task. & \href{https://doi.org/10.1016/j.neuroimage.2014.12.015}{FitzGerald, Moran et al.~(2015)} \cite{Fitzgerald2015a} and \href{https://doi.org/10.1162/NECO_a_00699}{FitzGerald, Schwartenbeck et al.~(2015)} \cite{Fitzgerald2015b} \\\hline
Psychopathology & Simulation of addictive choice behaviour. & Schwartenbeck, \href{https://doi.org/10.1016/j.mehy.2014.12.007}{FitzGerald, Mathys, Dolan, Wurst et al.~(2015)} \cite{Schwartenbeck2015a} \\\hline
Dopamine & The precision of beliefs about policies provides a plausible description of dopaminergic discharges. & \href{https://doi.org/10.1098/rstb.2013.0481}{Friston et al.~(2014)} \cite{Friston2014} and \href{https://doi.org/10.3389/fncom.2015.00136}{FitzGerald, Dolan et al.~(2015)} \cite{Fitzgerald2015c} \\\hline
Functional magnetic resonance imaging & Empirical prediction and validation of dopaminergic discharges. & \href{https://doi.org/10.1093/cercor/bhu159}{Schwartenbeck, FitzGerald, Mathys, Dolan, Friston (2015)} \cite{Schwartenbeck2014} \\\hline
Maximal utility theory & Evidence in favor of surprise minimization as opposed to utility maximization in human decision-making. & \href{https://doi.org/10.1038/srep16575}{Schwartenbeck, FitzGerald, Mathys, Dolan, Kronbichler et al.~(2015)} \cite{Schwartenbeck2015b} \\\hline
Social cognition & Examining the effect of prior preferences on interpersonal inference. & \href{https://doi.org/10.3389/fnhum.2014.00160}{Moutoussis et al.~(2014)} \cite{Moutoussis2014} \\\hline
Exploration-exploitation dilemma & Casting behavior as expected free energy minimizing accounts for epistemic and pragmatic choices. & \href{https://doi.org/10.1080/17588928.2015.1020053}{Friston et al.~(2015)} \cite{Friston2015epistemic} \\\hline
Habit learning and action selection & Formulating learning as an inferential process and action selection as Bayesian model averaging. & \href{https://doi.org/10.1016/j.neubiorev.2016.06.022}{Friston et al.~(2016)} \cite{Friston2016} and \href{https://doi.org/10.3389/fnhum.2014.00457}{FitzGerald et al.~(2014)} \cite{Fitzgerald2014} \\\hline
Scene construction and anatomy of time & Mean-field approximation for multi-factorial hidden states, enabling high dimensional representations of the environment. & \href{https://doi.org/10.1016/j.tics.2016.05.001}{Friston and Buzsáki (2016)} \cite{Friston2016f} and \href{https://doi.org/10.3389/fncom.2016.00056}{Mirza et al.~(2016)} \cite{Mirza2016} \\\hline
Electrophysiological responses & Synthesizing various in-silico neurophysiological responses via a gradient descent on free energy. E.g., place-cell activity, mismatch negativity, phase-precession, theta sequences, theta–gamma coupling and dopaminergic discharges. & \href{https://doi.org/10.1162/NECO_a_00912}{Friston, FitzGerald et al.~(2017)} \cite{Friston2017process} \\\hline
Structure learning, curiosity and insight & Simulation of artificial curiosity and abstract rule learning. Structure learning via Bayesian model reduction. & \href{https://dl.acm.org/doi/abs/10.1162/neco_a_00999}{Friston, Lin et al.~(2017)} \cite{Friston2017curiosity} \\\hline
Hierarchical temporal representations & Generalization to hierarchical generative models with deep temporal structure and simulation of reading. & \href{https://doi.org/10.1016/j.neubiorev.2018.04.004}{Friston et al.~(2018b)} \cite{Friston2018deep} and \href{https://doi.org/10.1038/s41598-017-15249-0}{Parr and Friston (2017b)} \cite{Parr2017} \\\hline
Computational neuropsychology & Simulation of visual neglect, hallucinations, and prefrontal syndromes under alternative pathological priors. & \href{https://doi.org/10.1162/cpsy_a_00022}{Benrimoh et al.~(2018)} \cite{Benrimoh2018}, \href{https://doi.org/10.3389/fnint.2018.00039}{Parr, Benrimoh et al.~(2018)} \cite{Parr2018precision}, \href{https://doi.org/10.1093/cercor/bhx316}{Parr and Friston (2018c)} \cite{Parr2018}, \href{https://doi.org/10.3389/fnhum.2018.00061}{Parr, Rees et al.~(2018)} \cite{Parr2018comp} and \href{https://doi.org/10.1093/cercor/bhz118}{Parr, Rikhye et al.~(2019)} \cite{Parr2020prefrontal} \\\hline
Neuromodulation & Use of precision parameters to manipulate exploration during saccadic searches; associating uncertainty with cholinergic and noradrenergic systems. & \href{https://doi.org/10.1098/rsif.2017.0376}{Parr and Friston (2017a)} \cite{Parr2017uncertainty}, \href{https://doi.org/10.1007/s00213-019-05240-0}{Parr and Friston (2019)} \cite{Parr2019pharm}, \href{https://doi.org/10.1371/journal.pcbi.1006267}{Sales et al.~(2019)} \cite{Sales2019} and \href{https://doi.org/10.1371/journal.pcbi.1007126}{Vincent et al.~(2019)} \cite{Vincent2019} \\\hline
Decisions to movements & Mixed generative models combining discrete and continuous states to implement decisions through movement. & \href{https://doi.org/10.1162/netn_a_00018}{Friston, Parr et al.~(2017)} \cite{Friston2017a} and \href{https://doi.org/10.1162/neco_a_01102}{Parr and Friston (2018d)} \cite{Parr2018discrete} \\\hline
Planning, navigation and niche construction & Agent induced changes in environment (generative process); decomposition of goals into subgoals. & \href{https://doi.org/10.1016/j.jtbi.2018.07.002}{Bruineberg et al.~(2018)} \cite{Bruineberg2018}, \href{https://doi.org/10.1098/rsif.2017.0685}{Constant et al.~(2018)} \cite{Constant2018} and \href{https://doi.org/10.1007/s00422-018-0753-2}{Kaplan and Friston (2018a)} \cite{Kaplan2018} \\\hline
Atari games & Active inference compares favorably to reinforcement learning in the game of Doom. & \href{https://doi.org/10.1016/j.bpsc.2018.06.010}{Cullen et al.~(2018)} \cite{Cullen2018} \\\hline
Machine learning & Scaling active inference to more complex machine learning problems. & \href{https://doi.org/10.1109/IJCNN48605.2020.9207382}{Tschantz et al.~(2019)} \cite{Tschantz2020} \\\hline
\end{longtable}}
\end{center}

\begin{sloppypar}
\printbibliography[title={Supplemental References}]
\end{sloppypar}
\end{refsection}

\end{document}